\documentclass[10pt,final,journal]{IEEEtran}
\usepackage{pdfpages}
\pdfoutput=1
\usepackage{amsmath,epsfig,amsfonts,multirow}
\usepackage{tabularx}
\usepackage{times}
\usepackage{graphicx}
\usepackage{amssymb}
\usepackage{algorithm}
\usepackage{algorithmic}
\usepackage{placeins}
\usepackage{pbox}
\usepackage{cite}

\newcommand{\argmin}{\mathop{\mathrm{arg\,min}}}
        
\begin{document}

\title{Scalable image coding based on epitomes}

\author{Martin~Alain,
        Christine~Guillemot,~\IEEEmembership{Fellow,~IEEE,}
        Dominique~Thoreau,
        and~Philippe~Guillotel}

\markboth{Preprint submitted to IEEE Trans. on Image Processing, Jun. 2016}%
{}

\maketitle

\begin{abstract}
	
In this paper, we propose a novel scheme for scalable image coding based on the concept of epitome.
An epitome can be seen as a factorized representation of an image.
Focusing on spatial scalability, the enhancement layer of the proposed scheme contains only the epitome of the input image.
The pixels of the enhancement layer not contained in the epitome are then restored using two approaches inspired from local learning-based super-resolution methods.
In the first method, a locally linear embedding model is learned on base layer patches and then applied to the corresponding epitome patches to reconstruct the enhancement layer.
The second approach learns linear mappings between pairs of co-located base layer and epitome patches.
Experiments have shown that significant improvement of the rate-distortion performances can be achieved compared to an SHVC reference.

\end{abstract}

\begin{IEEEkeywords}
Epitome, super-resolution, scalable image coding, SHVC
\end{IEEEkeywords}

\IEEEpeerreviewmaketitle

\section{Introduction}

The latest HEVC standard \cite{Sullivan2012b} is among the most efficient codec for image and video compression \cite{Ohm2012}.
However, the ever increasing spatial and/or temporal resolution, bit depth, or color gamut of modern images and videos, coupled with the heterogeneity of the distribution networks, calls for scalable coding solutions.
Thus, a scalable extension of HEVC named SHVC was developed \cite{Sullivan2013a,Ye2014}, which can encode enhancement layers with the scalability features mentioned above by using the appropriate inter-layer processing.
Experiments demonstrate that SHVC outperforms simulcast as well as the previous scalable standard SVC \cite{Kessentini2015}.
In this paper, we focus on spatial scalability and we propose a novel scalable coding scheme based on the concept of epitome, first introduced in \cite{Jojic2003,Cheung2008}. The epitome in \cite{Jojic2003} is defined as patch-based appearance and shape probability models learned from the image patches.
The authors have shown that these probability models, together with appropriate inference algorithms, are useful for content analysis, inpainting or super-resolution.
A second form of epitome has been introduced in \cite{Wang2008} which can be seen as a summary of the image.
This epitome is constructed by searching for self-similarities within the image using methods such as the KLT tracking algorithm.
This type of epitome has been used for still image compression in \cite{Cherigui2011} where the authors propose a rate-distortion optimized epitome construction method.
The image is represented by its epitome together with a transformation map as well as a reconstruction residue.
A novel image coding architecture has also been described in \cite{Cherigui2014} which, instead of the classical block processing in a raster scan order, inpaints the epitome with in-loop residue coding.

We describe in this paper a novel spatially scalable image coding scheme in which the enhancement layer is only composed of the input image epitome.
This factorized representation of the image is then used by the decoder to reconstruct the entire enhancement layer using single-image super-resolution (SR) techniques.
Single-image SR methods can be broadly classified into two main categories: the interpolation-based methods \cite{Li2001,Tappen2003,Fattal2007} and the example-based methods \cite{Freeman2000,Freeman2002,HongChang2004,Fan2007,Glasner2009,Yang2010a,Yang2013,Bevilacqua2014,Zhang2015} which we consider here, focusing on two different techniques based on neighbor embedding \cite{HongChang2004} and linear mappings \cite{Yang2013}. 
The epitome patches transmitted in the enhancement layer (EL) and the corresponding base layer (BL) patches form a dictionary of pairs of high-resolution and low-resolution  patches.

The first method based on neighbor embedding assumes that the BL and EL patches lie on two low and high resolution manifolds which share a similar local geometrical structure.
In order to reconstruct an EL patch not belonging to the epitome, a local model of the corresponding BL patch is learned as a weighted combination of its nearest neighbors in the dictionary.
The restored EL patch is then obtained by applying this weighted combination to the corresponding EL patches in the dictionary.
The second approach based on linear mappings rely on a similar assumption, but directly models a projection function between BL patches and the corresponding EL patches in the dictionary.
The projection function is learned using multivariate regression and is then applied to the current BL patch in order to obtain its restored EL version.
This super-resolution step reconstructs the full enhancement layer while we only transmit the epitome.
The proposed scheme thus allows reaching significant bit-rate reduction compared to traditional scalable coding schemes such as SHVC.

This paper is organized as follows.
In section \ref{sota:eptm} we review the background on epitomic models.
Section \ref{sec:eptm_svc} describes the proposed scheme, the epitome generation and encoding at the encoder side, and the epitome-based restoration at the decoder side.
Finally, we present in section \ref{sec:results} the results compared with SHVC.

\section{Background on epitomes}
\label{sota:eptm}

The concept of epitome was first introduced by N. Jojic and V. Cheung in \cite{Jojic2003,Cheung2008}.
It is defined as the condensed representation (meaning its size is only a fraction of the original size) of an image signal containing the essence of the textural properties of this image.
This original epitomic model is based on a patch-based probabilistic approach.
It was shown to be of high ``completeness'' in \cite{Simakov2008}, but introduces undesired visual artifacts, which is defined as a lack of ``coherence''.
In fact since the model is learned by compiling patches drawn from the input image, patches that were not in the input image can appear in the epitome. 
The original epitomic model was also extended into a so-called Image-Signature-Dictionary (ISD) optimized for sparse representations \cite{Aharon2008}.

The aforementioned epitomic models have been successfully applied to segmentation, de-noising, recognition, indexing or texture synthesis.
The model of \cite{Jojic2003,Cheung2008} was also used in \cite{Wang2012} for intra coding. 
However, this epitomic model is not designed for image coding applications, and thus have to be coded losslessly, which limits the compression performances.

\begin{figure}[t]
	\begin{tabular}{cc}
	Epitome	& Reconstructed image \\
	34 \% of input image & PSNR = 35.53 dB \\
	\includegraphics[width=4cm]{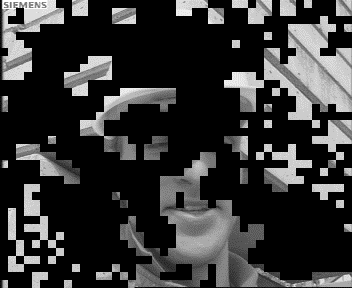} &
	\includegraphics[width=4cm]{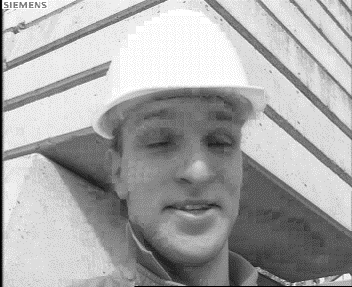} \\
	\end{tabular}
	\caption{Epitome of a Foreman frame (left) and the corresponding reconstruction (right).}
\label{fig:Formean_eptm}
\end{figure}

Thus, the work presented in this paper is derived from the approach introduced in \cite{Cherigui2011}.
This epitomic model is dedicated to image coding, and was inspired by the factorized image representation of Wang et al \cite{Wang2008}.
In this approach the input image $I$ is factored in an epitome $E$, which is composed of disjoint texture pieces called epitome charts (see Fig. \ref{fig:Formean_eptm}).
The input image is divided into a regular grid of non-overlapping blocks $B_i$ (block-grid) and each block is reconstructed from an epitome patch.
A so-called assignation map links the patches from the epitome to the reconstructed image blocks.
This epitomic model is obtained through a two-step procedure which first searches for the self-similarities within the input image, and then iteratively grows the epitome charts.
The second step for creating the epitome charts is notably based on a rate-distortion optimization (RDO) criterion, which minimizes the distortion between the input and the reconstructed image together with the rate of the epitome, evaluated as its number of pixels.

A still image coding scheme based on this epitomic model is also described in \cite{Cherigui2011}, where the epitome and its associated assignation map are encoded.
The reconstructed image can thus be used as a predictor, and the corresponding prediction residue is further encoded.
The results show that the scheme is efficient against H.264 Intra.
However, the coding performances of the assignation map are limited, which reduces the overall rate-distortion (RD) gains.

A novel coding scheme was thus proposed in \cite{Cherigui2014}, where only the epitome is encoded.
The blocks not belonging to the epitome are then predicted in an inpainting fashion, together with an in-loop encoding of the residue.
The prediction tools notably include efficient template-based neighbor embedding techniques such as the Locally Linear Embedding (LLE) \cite{Roweis2000,Turkan2012}.
The results show that significant bit-rate savings are achieved with respect to H.264 Intra.

In the next section, we describe the proposed scheme, which can be seen as an extension of the latter work to scalable coding.

\section{Epitomic enhancement layer for scalable image coding}
\label{sec:eptm_svc}

\begin{figure*}[ht]
	\centering
	\includegraphics[width=13cm,trim={0cm 0cm 8.5cm 10cm},clip]{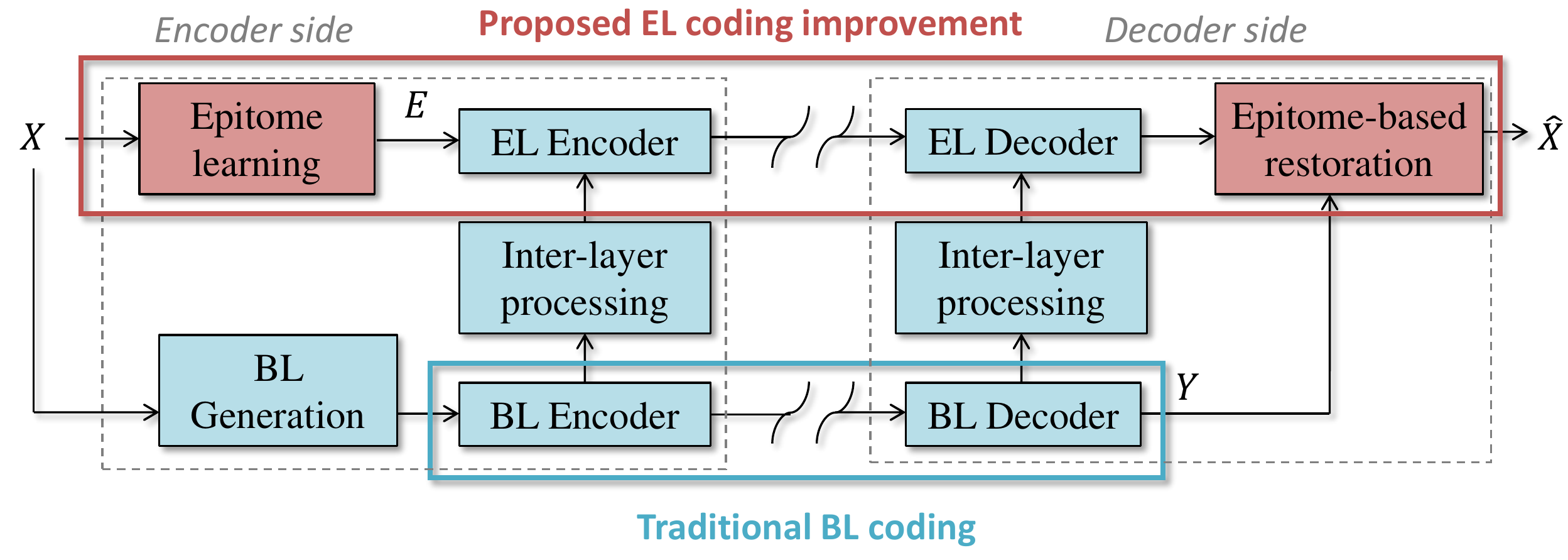}
	\caption{Proposed scheme for scalable image coding. At the encoder side, an epitome of the input image is generated, and encoded as the enhancement layer. At the decoder side, the enhancement layer patches not contained in the epitome are reconstructed from the base layer.}
	\label{fig:proposed_scheme}
\end{figure*}

In this section, we describe a scalable coding scheme in which the enhancement layer consists in an epitome of the input image.
Consequently, at the decoder side, the EL patches not contained in the epitome are missing, but the corresponding BL patches are known.
We thus propose to restore the full enhancement layer by taking advantage of the known representative texture patches available in the EL epitome charts.
The proposed scheme is shown in Fig. \ref{fig:proposed_scheme}.

We first summarize below the two-step procedure for constructing the epitome, as well as its encoding process.
Then, we explain in details how to perform the restoration, using local learning-based techniques. 

\subsection{Epitome generation and encoding}

The epitome generation method used in this paper is overall derived from \cite{Cherigui2011}, and consists first in a self-similarities search step followed by an epitome charts creation step.
However, the self-similarities search step is here performed based on a fast two-step method that we proposed in \cite{Alain2014}.
Moreover, we choose here to use an error minimization criterion for the epitome chart creation instead of the RDO criterion of \cite{Cherigui2011}, as we noticed that in practice this RDO criterion has a limited impact on our application.

\begin{figure}[t]
	\begin{center}
	\includegraphics[width=8cm,trim={0cm 0cm 7cm 9cm},clip]{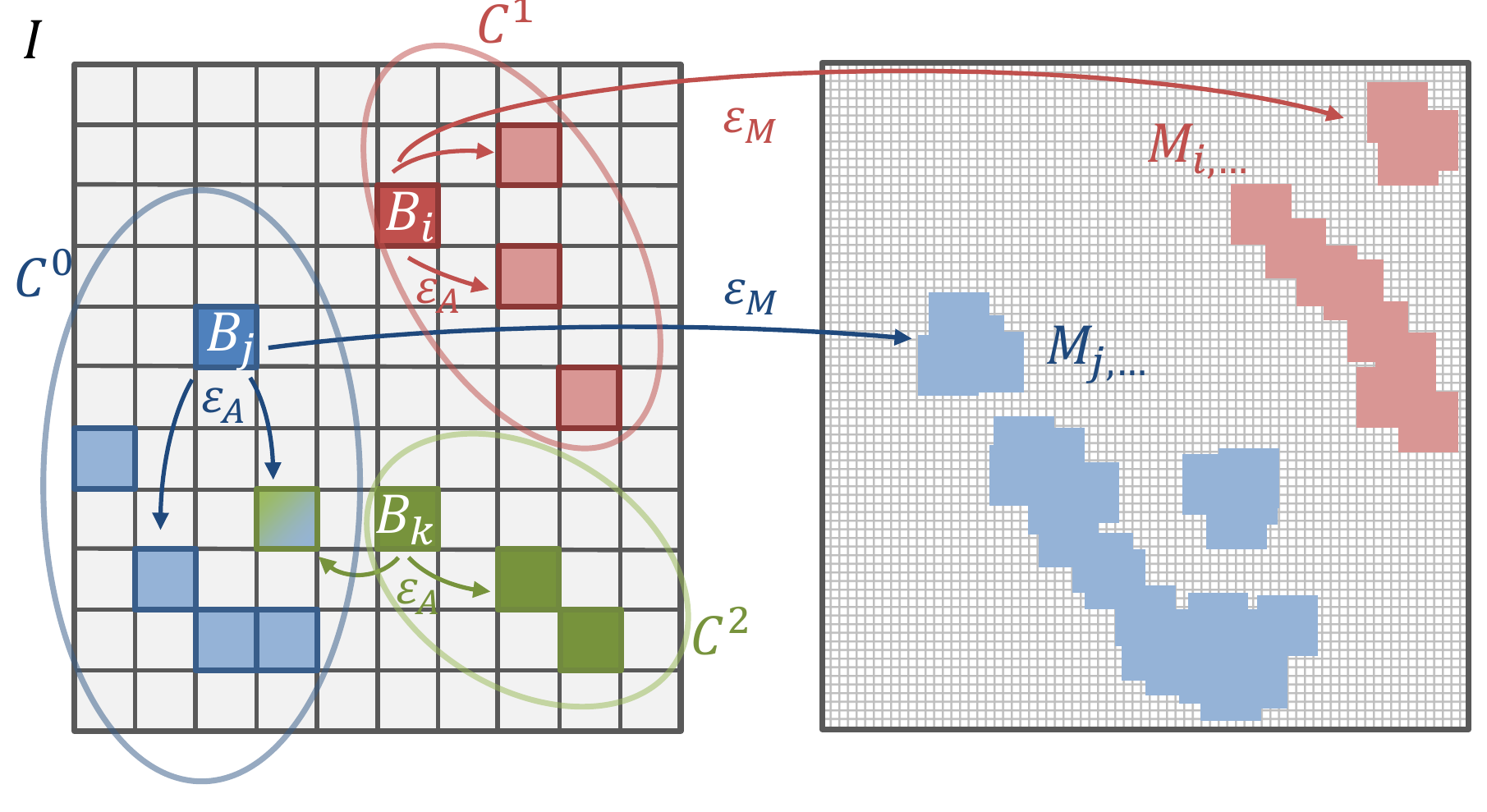}
	\end{center}
	\caption{Two steps clustering-based self-similarities search.}
\label{fig:eptm_cluster}
\end{figure}

\begin{figure}[t]
	\begin{center}
	\includegraphics[width=8cm,trim={0cm 0cm 7cm 9.5cm},clip]{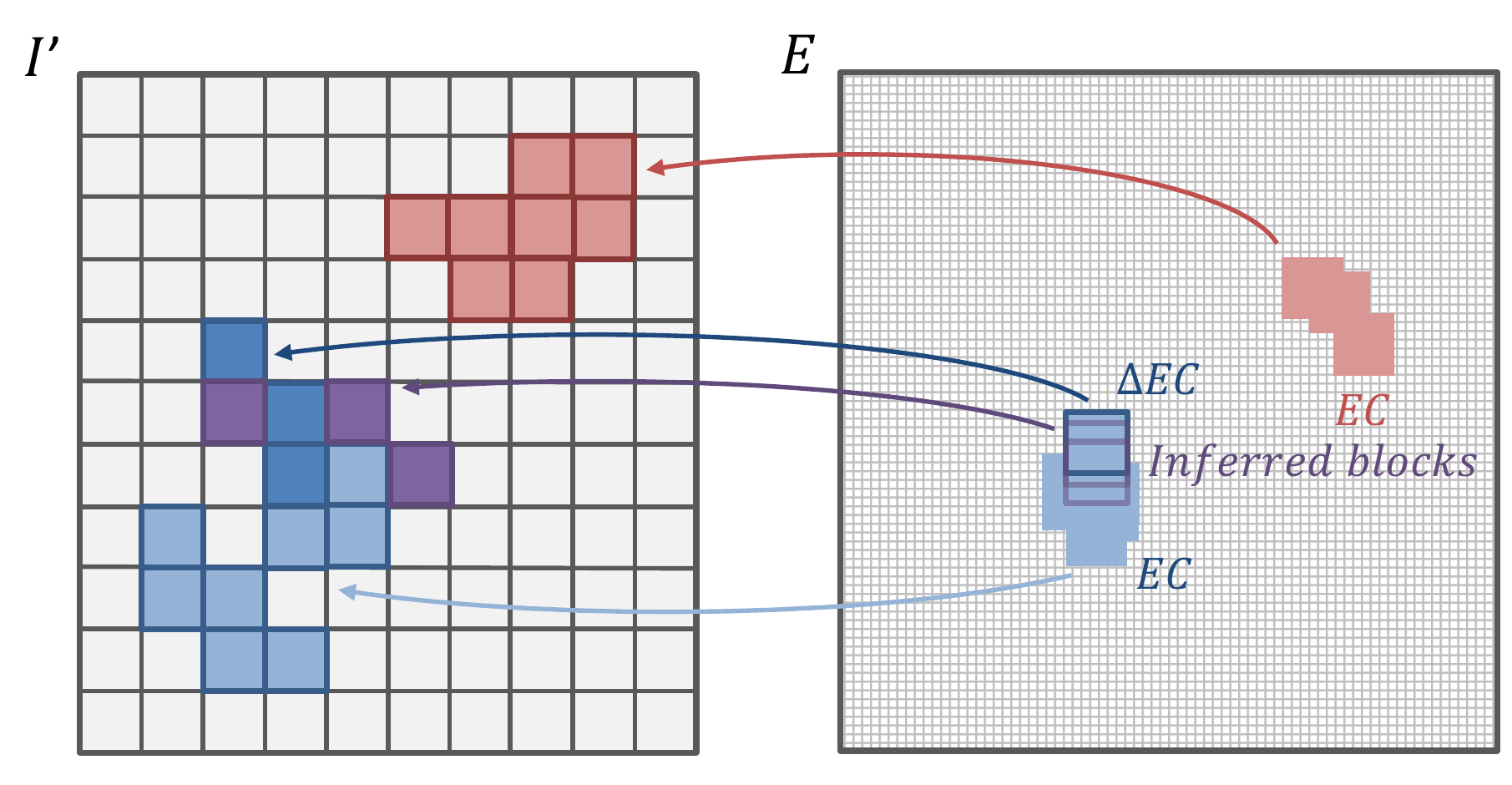}
	\end{center}
	\caption{Epitome chart extension process with inferred blocks.}
\label{fig:eptm_rec_im}
\end{figure}

\subsubsection{Self-similarities search}
\label{sota:self_sim}

The goal of this step is to find for each block $B_i \in I$ a list of matching patches $ML(B_i) = \lbrace M_{i,0}, M_{i,1}, ... \rbrace$, such that the mean square error (MSE) between a block and its matching patches is below a matching threshold $\varepsilon_M$. 
This parameter eventually determines the size of the epitome, and several values are considered in the experiments (see Table \ref{tab:eptm_size}).
In this paper, the lists of matching patches are obtained through a two steps clustering-based approach illustrated in Fig. \ref{fig:eptm_cluster}.
The first step consists in grouping together similar blocks into clusters, so that the distance from a block to the centroid of its cluster is below an assignation threshold $\varepsilon_A$.
In practice, this threshold is set to $\varepsilon_A = 0.5 * \varepsilon_M$.
In the second step, a list of matching patches is computed for each cluster by finding patches whose MSE is inferior to $\varepsilon_M$ with respect to the block closest to the cluster centroid.
This list of matching patches is then assigned to all the blocks in the cluster.

From all the lists of matching patches $ML(B_i)$, we can then compute reverse lists $RL(M_{j})=\lbrace B_j,B_l, ... \rbrace$ which indicate the set of image blocks that can be represented by each matching patch.
Next, we describe how these lists are used to build the epitome charts.

\subsubsection{Epitome charts creation}
\label{sota:eptm_gen}

The epitome charts are iteratively grown, and both the initialization and the iterative extension of an epitome chart are based on a criterion which minimizes the error between the input image $I$ and the reconstructed image $I'$.

Formally, we denote $\Delta EC_m, m=0,...,M-1$ a set of candidate regions to add to the epitome, where $M$ is the number of candidates.
When initializing a new epitome chart, a valid candidate region is a matching patch which is not yet in an epitome chart and is spatially disconnected from any existing epitome chart.
On the contrary, when extending an epitome chart $EC$, a valid candidate region is a matching patch which is not yet in an epitome chart and overlaps with $EC$.
The actual region added to the  epitome $\Delta EC_{opt}$ is obtained by minimizing the following criterion:

\begin{equation}
\Delta EC_{opt} = \argmin_{\Delta EC_m}(MSE(I,I'_m))
\label{eq:ext_crit}
\end{equation}

\noindent where $I'_m$ is the reconstructed image when the candidate region $\Delta EC_m$ is added to the epitome, and the $MSE$ function computes the mean square error between 
$I'_m$ and the source image $I$.
The reconstructed image $I'_m$ comprises the blocks reconstructed from the existing epitome charts and the new blocks contained in the list $RL(\Delta EC_m)$.
During the extension of an epitome chart, additional reconstructed blocks can be obtained by considering the so-called inferred blocks, which are the potential matching patches that can overlap between the current chart $EC$ and the extension $\Delta EC_m$ (see Fig. \ref{fig:eptm_rec_im}).
Note that for the pixels of $I'_m$ which are not reconstructed, the MSE can not be computed directly.
In our implementation, we assign the maximal MSE value to these pixels.
This tends to favor the selection of a candidate region $\Delta EC_m$ which reconstructs large regions in $I'_m$, and thus speed up the epitome chart creation.

The extension of an epitome chart stops when no more valid candidate regions can be found.
A new epitome chart is then initialized at a new location. 
The global process stops when the whole image $I'$ is reconstructed.
Note that the epitome charts in $E$ are originally obtained at a pixel accuracy, but for coding purposes they are then padded to be aligned with the block structure of the encoder.

\subsubsection{Epitome encoding}

The epitomes are encoded with a scalable scheme as an enhancement layer.
The blocks not belonging to the epitome are directly copied from the decoded base layer, thus their rate-cost is practically non-existent.

\subsection{Epitome-based restoration}
\label{sec:eptm_SR}

\begin{figure*}[ht]
	\centering
	\includegraphics[width=18cm,trim={0cm 0cm 0cm 4cm},clip]{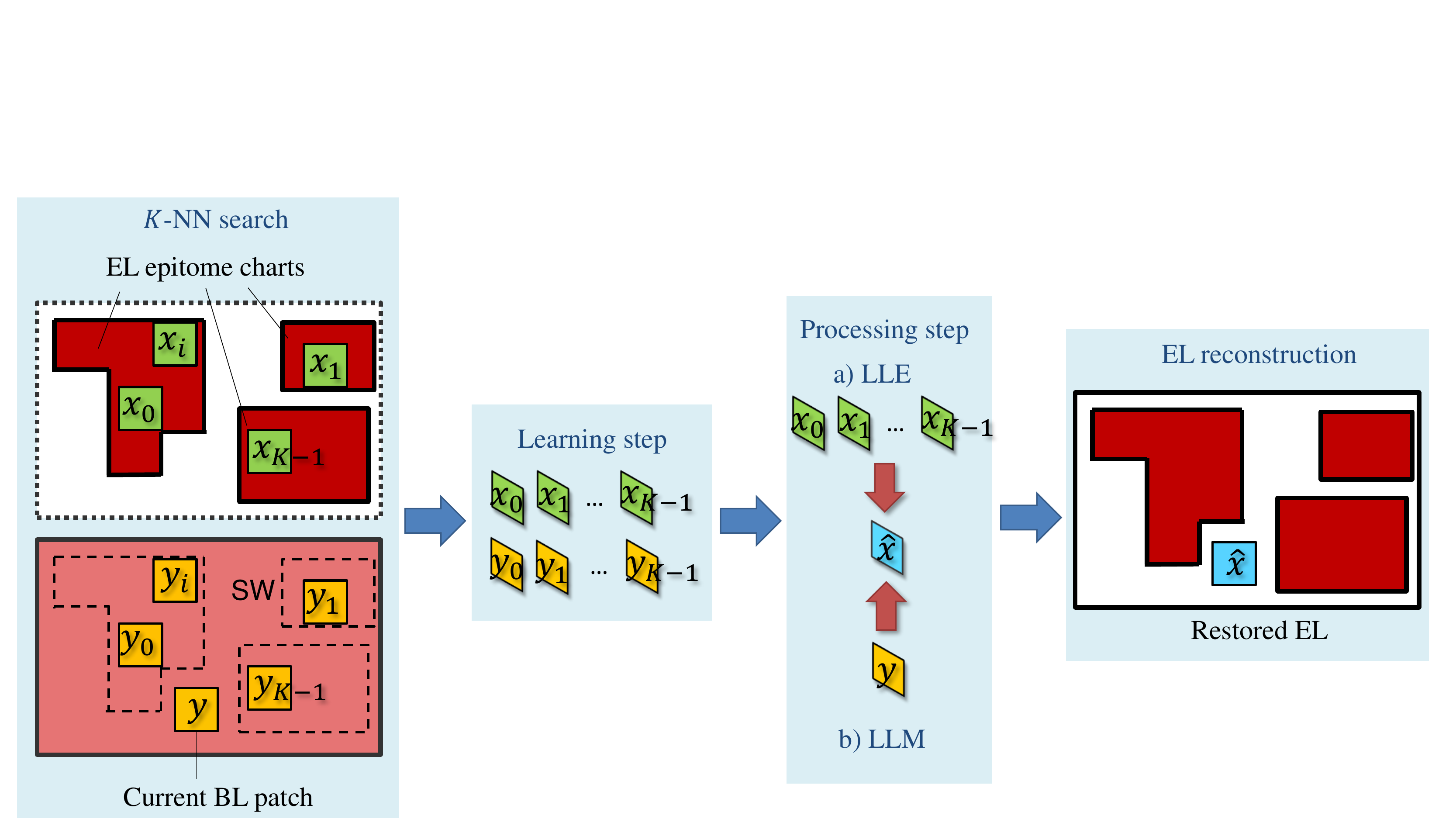}
	\caption{The $K$ nearest neighbors of the current BL patch are found in search windows (SW) corresponding to the epitome charts. The BL/EL pairs of patches can then be exploited to restore the current missing EL patch.}
	\label{fig:KNN_eptm}
\end{figure*}

The non-epitome part of the enhancement layer is processed by considering $N \times N$ overlapping patches, separated by a step of $s$ pixels in both rows and columns.
After restoration, when several estimates are obtained for a pixel, they are averaged in order to obtain the final estimate.
Note that before performing the restoration, the BL image is up-sampled to the resolution of the EL using the inter-layer processing filter.

The restoration methods described below are derived from local learning-based SR methods \cite{HongChang2004,Fan2007,Bevilacqua2014,Yang2013,Zhang2015}, and can be summarized in the three following steps: 
$K$-NN search, learning step, and processing step.
These steps are shown in Fig. \ref{fig:KNN_eptm}, and described in details below.

\subsubsection{K-NN search}
\label{sec:knn}

If we consider the current patch to be processed $y$, we first search for its $K$-NN BL patches, within search windows corresponding to the epitome charts locations (see Fig. \ref{fig:KNN_eptm}).
The $K$-NN BL patches $y_i, i = 1 \dots K$ are then stored in a matrix $\textbf{M}_y$ which contains in its columns the vectorized patches.
For each neighbor $y_i$, we have a corresponding EL patch $x_i$ in the epitome, which is stored in a matrix $\textbf{M}_x$.
We thus obtain BL/EL pairs of training patches.
In classical SR applications, the pairs of training patches are obtained from a dictionary, which construction is a critical step \cite{Freeman2000,Freeman2002,HongChang2004,Yang2010a,Zhang2015,Glasner2009,Bevilacqua2014}. 
Since here the patches in the epitome are representative of the full image, we can consider that they constitute a suitable dictionary to perform the local learning-based restoration.

Next, we present the learning and processing steps, which exploit the correlation between the pairs of training patches to perform learning-based restoration.
We describe two methods to restore the missing EL patches, inspired by SR techniques based on neighbor embedding (NE) \cite{HongChang2004,Fan2007,Bevilacqua2014} and linear  mappings \cite{Yang2013,Bevilacqua2014,Zhang2015}, but any other learning-based method could be included in the proposed scheme. 
Note that many SR methods can be improved using iterative back-projection \cite{Irani1991}, which enforces the high resolution reconstructed image to be consistent with the input low resolution image.
However, this technique will not be considered in the proposed scheme, as it tends to propagate quantization noise from the BL image to the EL reconstruction.

\subsubsection{Epitome-based Locally Linear Embedding}
\label{sec:eptm_LLE}

\begin{figure}[ht]
	\centering
	\footnotesize
	\begin{tabular}{cc}
	\includegraphics[width=3.5cm,trim={0cm 14cm 28cm 0cm},clip]{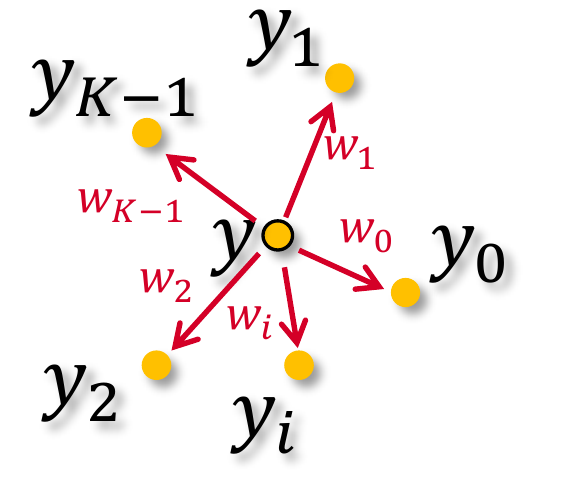} &
	\includegraphics[width=3.5cm,trim={5.5cm 14cm 23cm 0cm},clip]{learning_based_restoration.pdf} \\
	\parbox{4cm}{a) Learning: estimate the current BL patch as linear combination of its $K$-NN BL patches.} & 
	\parbox{4cm}{b) Processing: Apply the weights of the linear combination learned previously on the $K$-NN corresponding EL patches to obtain the restored EL patch.} \\
	\end{tabular}
	\caption{Two steps local learning-based restoration based on LLE.}
	\label{fig:E-LLE}
\end{figure}

First, we describe a method relying on LLE, denoted ``epitome-based Locally Linear Embedding'' (E-LLE).
Similarly to other NE-based restoration techniques, we assume that the local geometry of the manifolds in which lie the BL and EL patches is similar (see Fig. \ref{fig:E-LLE}).
Using LLE, we first learn the linear combination of the $K$-NN BL patches which best approximate the current patch, and then apply this linear combination to the corresponding EL patches in order to obtain a good estimate of the missing EL patch.

Let $W$ be the vector containing the combination weights $w_i, i = 1 \dots K$.
The weights are obtained by solving the following equation:

\begin{equation}
\min_{W} ||y-{\bf M}_y W ||^2_2~\text{s.t.}~\sum\limits_{i=1}^K w_i = 1
\end{equation}

\noindent 
The weights vector $W$ is computed as:

\begin{equation}
W = \frac{{{{\bf D}^{ - 1}}{\bf 1}}}{{{{\bf 1}^{\rm T}}{{\bf D}^{ - 1}}{\bf 1}}}.
\end{equation}

\noindent The term ${\bf D}$ denotes the local covariance matrix (\textit{i.e.}, in reference to $y$) of the $K$-NN stacked in ${\bf M}_y$, and $\bf 1$ is the column vector of ones.
In practice, instead of an explicit inversion of the matrix ${\bf D}$, the linear system of equations ${\bf D} W=\bf 1$ is solved, then the weights are rescaled so that they sum to one.

The restored EL patch is finally obtained as:

\begin{equation}
\hat{x} = {\bf M}_x W.
\end{equation}

In practice, several versions of this method can be derived, e.g. by using another NE-based technique such as non-negative matrix factorization \cite{Bevilacqua2012}, or by adapting the weights computation as in the non-local mean algorithm (exponential weights) \cite{Buades2005}. 
However, with such methods the weights are only computed based on the BL patches.
In the next section, we propose a method which aims at better exploiting the correlation between the pairs of training patches, based on linear regression.

\subsubsection{Epitome-based Local Linear Mapping}
\label{sec:eptm_LLM}

\begin{figure}[ht]
	\centering
	\footnotesize
	\begin{tabular}{c}
	\includegraphics[width=7cm,trim={22cm 5.5cm 0cm 7cm},clip]{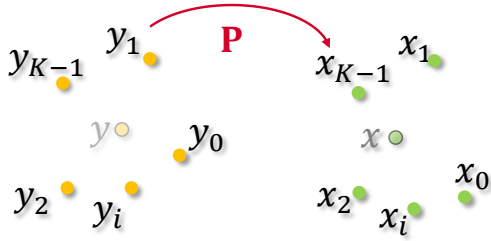} \\
	\parbox{8cm}{a) Learning: a mapping function is learned between the pairs of BL/EL patches using multivariate linear regression.} \\
	\includegraphics[width=7cm,trim={22cm 0cm 0cm 13.5cm},clip]{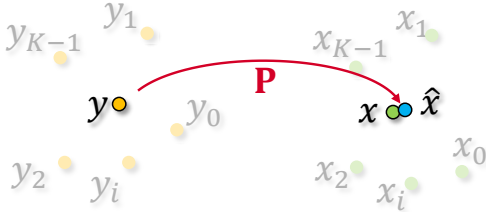} \\
	\parbox{8cm}{b) Processing: the function learned previously is applied on the current BL patch in order to obtain the restored EL patch.} \\
	\end{tabular}
	\caption{Two steps local learning-based restoration based on linear regression.}
	\label{fig:E-LLM}
\end{figure}

We describe here a method based on linear regression, that we denote ``epitome-based Local Linear Mapping (E-LLM)''.
We want to further exploit the correlation between the pairs of training patches by directly learning a function mapping the BL patches to the corresponding EL patches (see Fig. \ref{fig:E-LLM}).
This function can then be applied on the current patch to restore the EL patch.

The mapping function is learned using multivariate linear regression.
The problem is then to search for the function $\textbf{P}$ minimizing:

\begin{equation}
\mathbf{E}=||\textbf{M}_x^T - \textbf{M}_y^T \textbf{P}^T||^2
\label{eq:LM_LSform}
\end{equation}

\noindent which is of the form $||\textbf{Y}-\textbf{X}\textbf{B}||^2$ (corresponding to the linear regression model $\textbf{Y}=\textbf{X}\textbf{B}+\textbf{E}$).
The minimization of Eq. (\ref{eq:LM_LSform}) gives the least squares estimator

\begin{equation}
\textbf{P} = \textbf{M}_x \textbf{M}_y^T (\textbf{M}_y \textbf{M}_y^T)^{ - 1}
\label{eq:LM_proj}
\end{equation}

We finally obtain the restored EL patch as:

\begin{equation}
\hat{x} = \textbf{P} y
\end{equation}

Now that we have formally defined the proposed methods, we study their performances in the next section.

\section{Simulations and results}
\label{sec:results}

\subsection{Experimental conditions}

The experiments are performed on the test images listed in Table \ref{tab:test_im}, obtained from the HEVC test sequences.
The base layer images are obtained by down sampling the input image with a factor $2 \times	 2$, using the SHVC down-sampling filter available with the SHM software (ver. 9.0) \cite{SHMSoft}.
The BL images are encoded with HEVC, using the HM software (ver. 15.0) \cite{HMSoft}.

We then use the SHM software (ver. 9.0) \cite{SHMSoft} to encode the corresponding enhancement layers.
Thanks to the hybrid codec scalability feature of SHVC, the decoded BL images are first up-sampled using the separable 8-tap SHVC filter $(-1, 4, -11, 40, 40, -11, 4, -1)$, and directly used as input to the SHM software.
Both layers are encoded with the following quantization steps: QP = 22, 27, 32, 37.

\begin{table}[ht]
	\footnotesize
  \centering
  \caption{Test images}
    \begin{tabular}{|c|c|c|c|} 
    \hline
	\textbf{Class}	& \textbf{Image}	& \textbf{Size}			& \textbf{Epitome}		\\   
					& 					& 						& \textbf{block size}	\\   
	\hline                                                                                    
    B				&	BasketballDrive &	$1920 \times 1056$		& 	$16 \times 16$	\\   
   	B				&	Cactus 			&	$1920 \times 1056$		& 	$16 \times 16$	\\   
    B				&	Ducks 			&	$1920 \times 1056$		& 	$16 \times 16$	\\   
    B				&	Kimono 			&	$1920 \times 1056$  	& 	$16 \times 16$	\\   
   	B				&	ParkScene 		&	$1920 \times 1056$		& 	$16 \times 16$	\\   
    B				&	Tennis 			&	$1920 \times 1056$		& 	$16 \times 16$	\\   
    B				&	Terrace 		&	$1920 \times 1056$		& 	$16 \times 16$	\\   
	\hline                                                                                    
    C				&	BasketballDrill &	$832 \times 480$		& 	$8 \times 8$	\\   
    C				&	Keiba 			&	$832 \times 480$		& 	$8 \times 8$	\\   
    C				&	Mall  			&	$832 \times 480$		& 	$8 \times 8$	\\   
    C				&	PartyScene 		&	$832 \times 480$		& 	$8 \times 8$	\\   
	\hline                                                                                    
	D				&	BasketballPass 	&	$416 \times 240$		& 	$8 \times 8$	\\   
    D				&	BlowingBubbles 	&	$416 \times 240$		& 	$8 \times 8$	\\   
    D				&	RaceHorses 		&	$416 \times 240$		& 	$8 \times 8$	\\   
    D				&	Square 			&	$416 \times 240$		& 	$8 \times 8$	\\   
	\hline                                                                                    
    E				&	City  			&	$1280 \times 704$		& 	$16 \times 16$	\\   
    \hline
    \end{tabular}%
  \label{tab:test_im}%
\end{table}%

For each input image, 3 to 4 matching threshold values $\varepsilon_M$ are selected in order to generate epitomes which sizes range from 30\% to 90\% of the input image sizes.
The threshold values to reach such sizes vary depending on the input image, and were manually selected.
The selected matching thresholds and corresponding epitome sizes are shown in Table \ref{tab:eptm_size}.

\begin{table}[ht]
	\footnotesize
  \centering
  \caption{Epitome sizes (as \% of input images)}
    \begin{tabular}{|l|c|c|c|c|c|c|}
    \hline
    				& \multicolumn{6}{c|}{\textbf{Matching threshold} $\varepsilon_M$} \\
    \textbf{Image}	& 9		& 16	& 25	& 49	& 100	& 225 	\\
    \hline
    BasketballDrive & 90.62 & 64.10 & 49.33 & 32.34 &       &  		\\
    Cactus 			&       & 79.85 & 71.24 & 60.66 & 48.33 &  		\\     
    Ducks 			&       &       &       & 89.63 & 77.41 & 48.28 \\
    Kimono 			& 90.13 & 75.53 & 59.36 & 35.34 &       &  		\\     
    ParkScene 		& 86.58 &       & 73.55 & 61.99 & 47.18 &  		\\     
    Tennis 			& 64.49 & 50.44 & 43.12 & 32.22 &       &  		\\     
    Terrace 		& 78.46 &       & 66.39 &       & 53.31 & 43.50 \\
    BasketballDrill &       &       & 87.05 & 59.94 & 42.63 & 28.53 \\
    Keiba 			&       & 93.59 &       & 81.28 & 63.53 & 40.77 \\
    Mall  			& 92.95 &       &       & 76.28 & 66.15 & 50.26 \\
    PartyScene 		&       & 94.82 &       & 81.12 & 67.56 & 49.13 \\
	BasketballPass 	& 77.76 & 66.60 & 56.41 & 42.31 &       &  		\\     
    BlowingBubbles 	&       &       & 87.56 & 73.33 & 58.85 & 36.92 \\
    RaceHorses 		& 91.03 &       & 79.23 &       & 58.14 & 36.67 \\
    Square 			& 80.77 & 71.41 &       & 61.09 &       & 48.72 \\
	City  			&       &       & 91.59 & 82.44 & 66.81 & 39.52 \\
	\hline
    \end{tabular}%
  \label{tab:eptm_size}%
\end{table}%

The post-processing is performed using $N \times N = 8 \times 8$ overlapping patches, with an overlapping step $s = 3$.
We set the number of nearest neighbors to $K = 20$.

\subsection{Rate-distortion performances}

We assess in this section the performances of the proposed scheme against the SHVC reference EL. 
The distortion is evaluated using the PSNR of the decoded EL, while the rate is calculated as the sum of both BL and EL rates.
The RD performances are computed using the Bjontegaard rate gains measure (BD-rates) \cite{G2011} over the four QP values.

\begin{figure}[t]
	\centering
	\includegraphics[width=9cm,trim={1.6cm 2.2cm 2cm 1.9cm},clip]{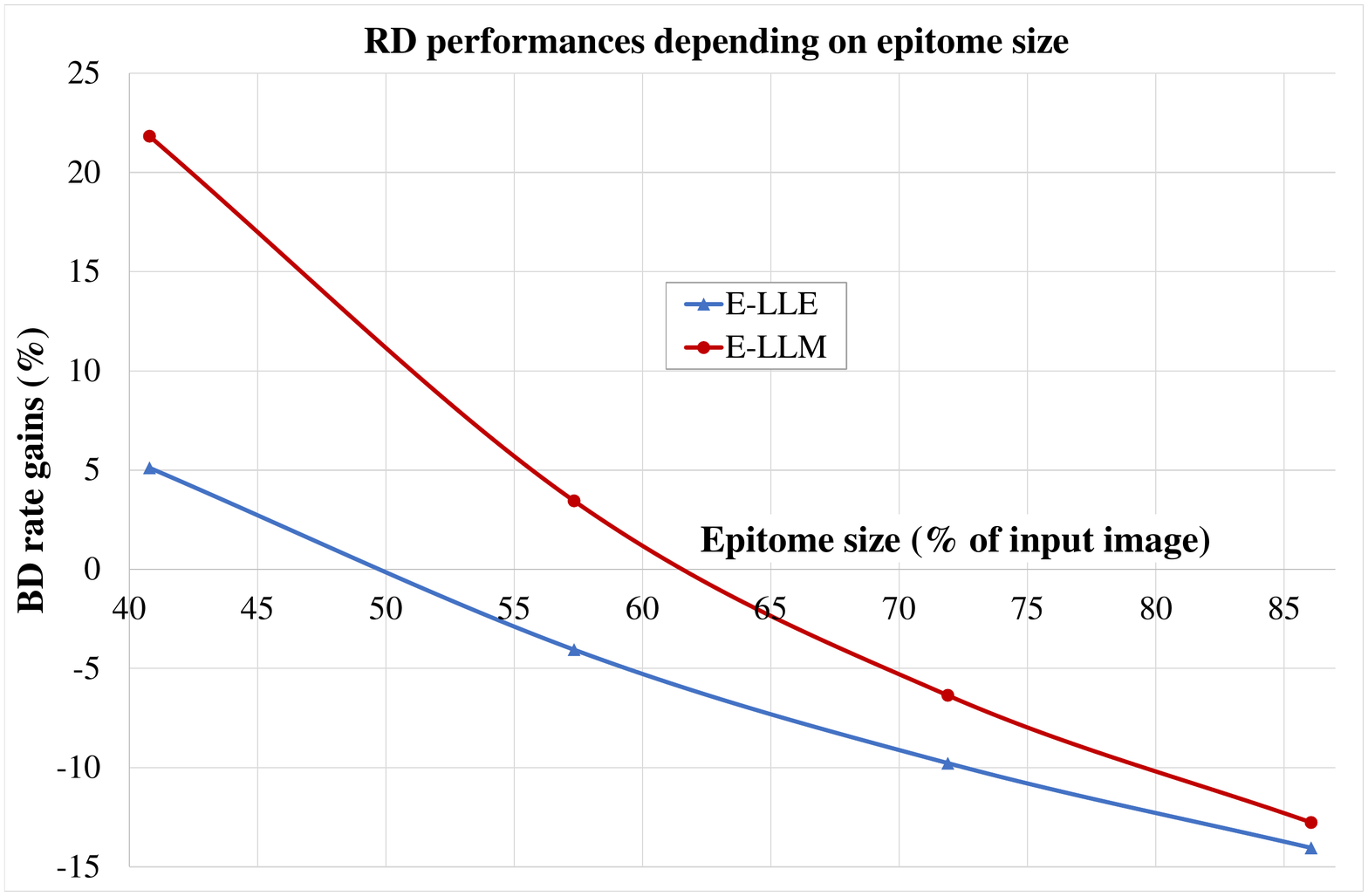}
	\caption{Average RD performances of the different restoration methods against SHVC depending on the epitome size.} 
	\label{fig:RD_perf_eptm_size}
\end{figure}

We show in Fig. \ref{fig:RD_perf_eptm_size} the BD-rates averaged over all sequences depending on the epitome size.
The complete results are given in Table \ref{tab:RD_perf}.
Overall, we can see that significant bit-rate reduction can be achieved compared to SHVC, up to about 15\% bit-rate reduction in average, and 20\% for images like BaskballDrive, Ducks, or Tennis.
The best performances are achieved with the biggest epitomes.
In fact, smaller epitomes provide a reduced set of BL/EL patches, while more BL patches need to be processed.
Eventually, the post-processing step can not effectively compensate for the quality loss.
Overall, the E-LLE performs better than the E-LLM method, however, for the best performances (biggest epitomes), both methods perform similarly.

\begin{figure}[t]
	\centering
	\includegraphics[width=9cm,trim={1.6cm 2.2cm 2cm 1.9cm},clip]{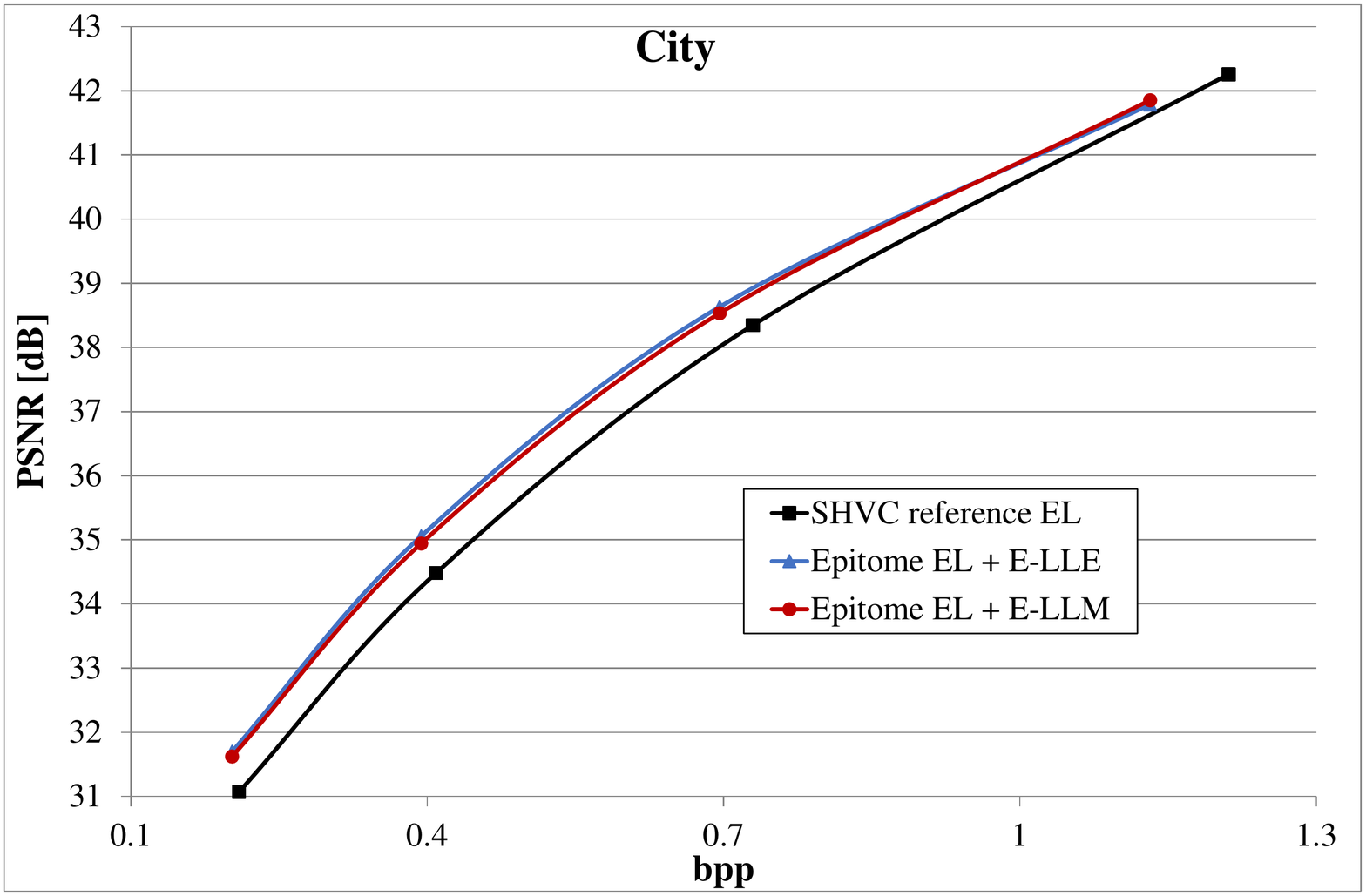}
	\caption{RD performances of the City image for both E-LLE and E-LLM methods, epitome size = 91.59\% of input image.} 
	\label{fig:RD_perf_City_PSNR}
\end{figure}

In order to better understand the performances of the proposed methods, we show in Fig. \ref{fig:RD_perf_City_PSNR} the RD curve of the City image for its best performances (biggest epitome), which behavior is representative of the set of test images.
We can see that at high bit-rates (QP=22), the bit-rate of the proposed scheme is especially reduced compared to the SHVC reference EL.
However, even with the proposed post-processing, we observe a loss of quality.
At low bit-rates (QP=37), the bit-rate of the proposed method is less reduced compared to the SHVC reference EL, but the post-processing yields a better quality.
This behavior explains the overall significant bit-rate reduction we can achieve with the proposed scheme.

\begin{figure}[t]
	\centering
	\includegraphics[width=9cm,trim={1.6cm 2.2cm 2cm 1.9cm},clip]{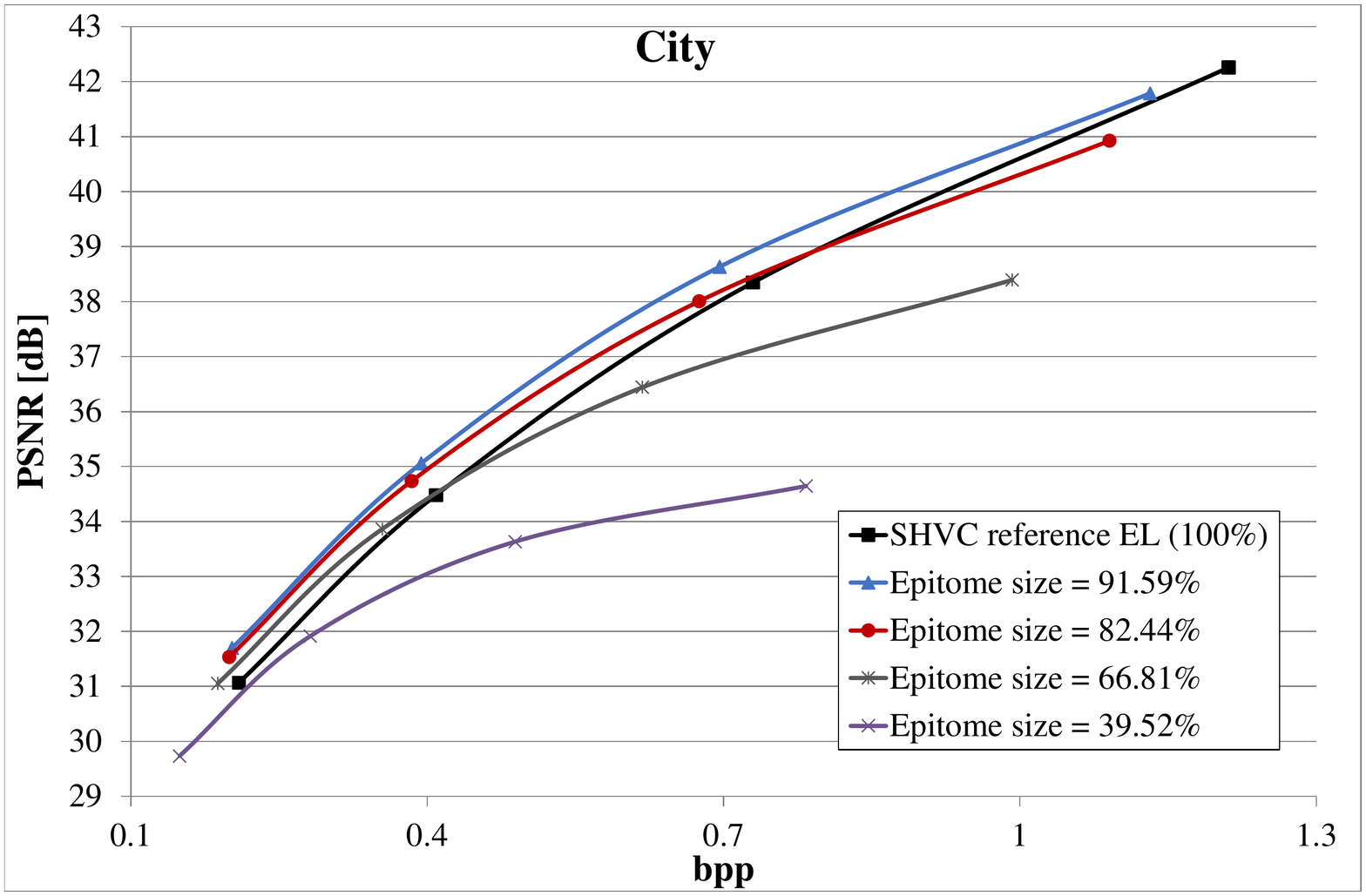}
	\caption{RD performances of the City image using the E-LLE method, with different epitome sizes.} 
	\label{fig:RD_perf_City_eptm_size}
\end{figure}

In addition, we show in Fig. \ref{fig:RD_perf_City_eptm_size} the RD curves of the LLE-based restoration for the PartyScene image depending on the epitome size.
We can see that for smaller epitome sizes, at high bit-rates (QP=22), even though the bit-rate is considerably reduced compared to the SHVC reference EL, the quality loss does not allow an improvement in the RD performances, which corroborates our previous analysis.

\begin{figure*}[htbp]
	\centering
	\begin{tabular}{cc}
	\multicolumn{2}{c}{Epitome EL before restoration} \\
	\multicolumn{2}{c}{\includegraphics[width=9cm,clip]{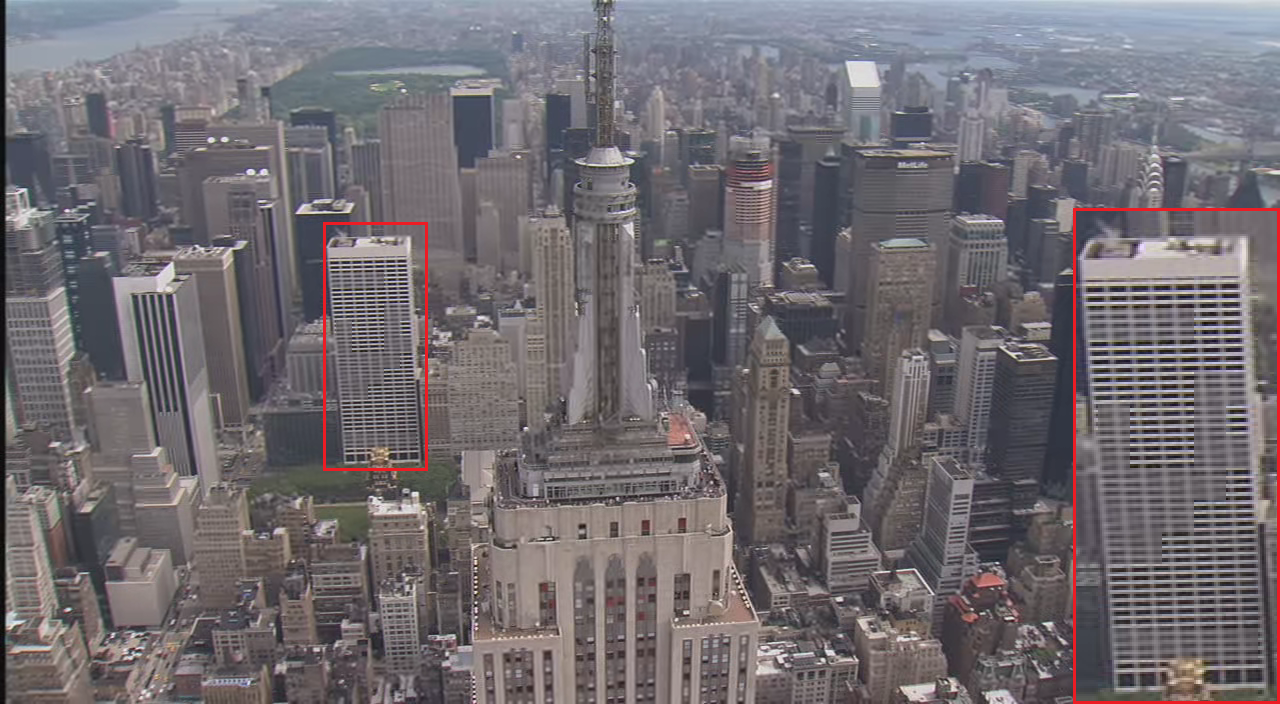}} \\
	Epitome EL + E-LLE		& Epitome EL + E-LLM			\\
	\includegraphics[width=9cm,clip]{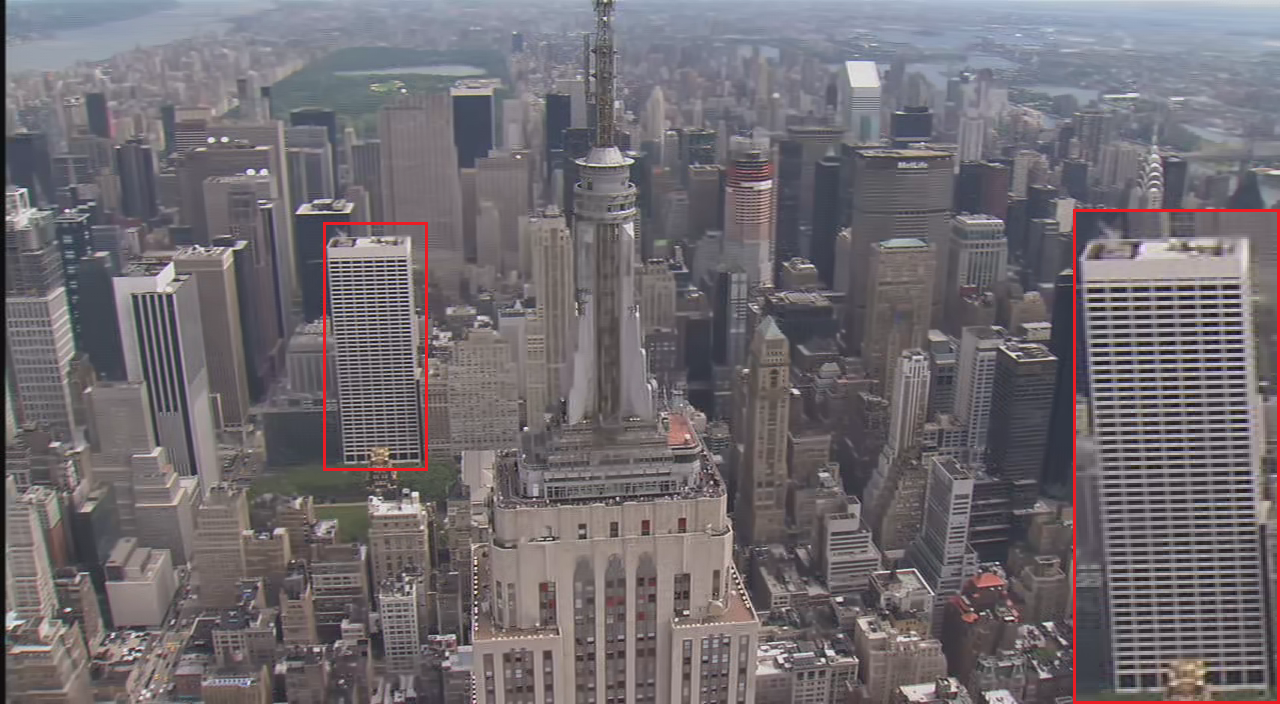} & 
	\includegraphics[width=9cm,clip]{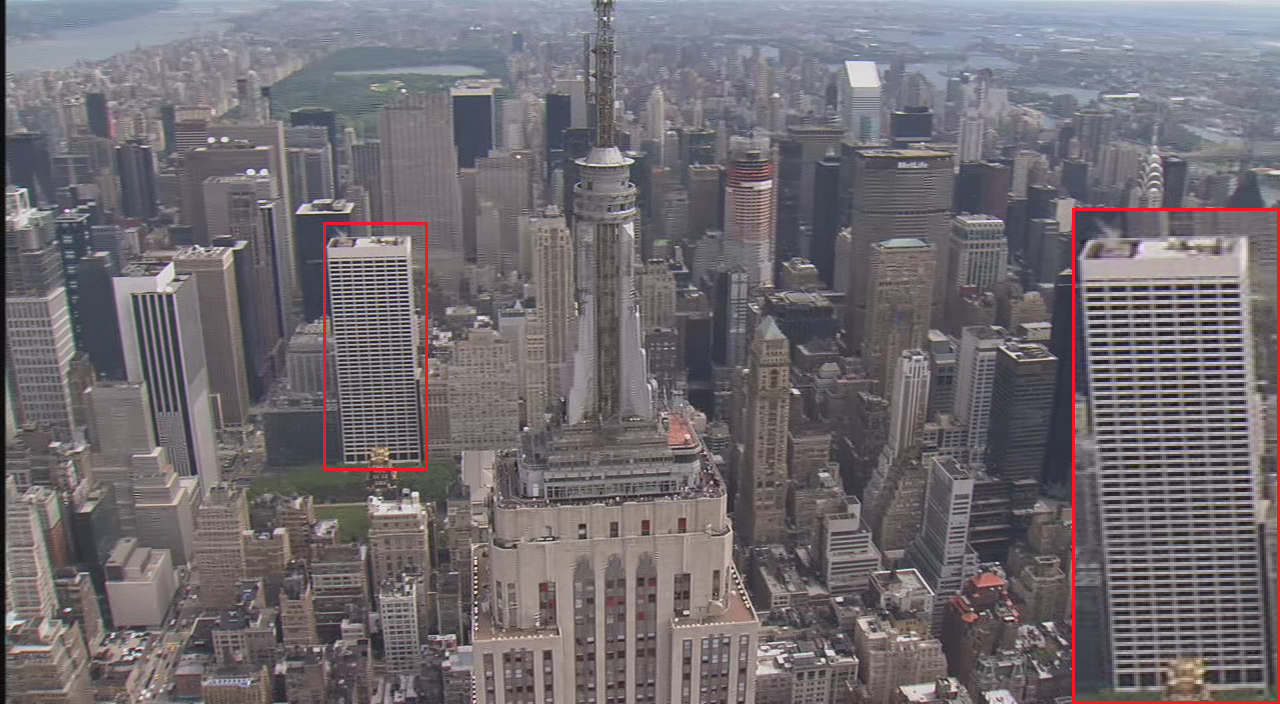} \\	
	\end{tabular}
	\caption{City enhancement layer encoded with QP=22. The epitome size is 39.52\% of the input image. Before restoration, we can clearly notice in the red rectangle the blocks not belonging to the epitome from their blurry aspect. The quality of these blocks is obviously improved after restoration.}
	\label{fig:City_example}
\end{figure*}

\begin{figure*}[htbp]
	\centering
	\begin{tabular}{cc}
	\multicolumn{2}{c}{Epitome EL before restoration} \\
	
	\multicolumn{2}{c}{\includegraphics[width=9cm,clip]{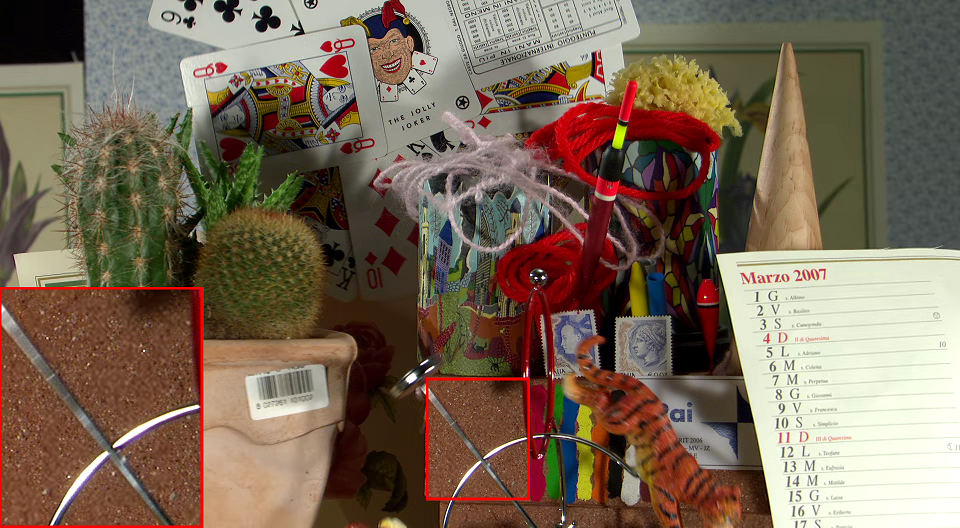}} \\
	Epitome EL + E-LLE & Epitome EL + E-LLM			\\
	\includegraphics[width=9cm,clip]{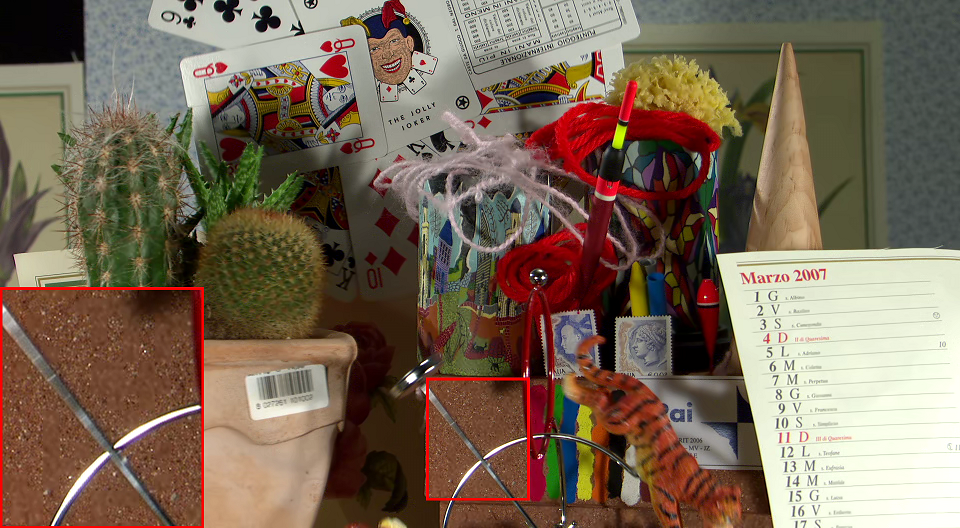} & 
	\includegraphics[width=9cm,clip]{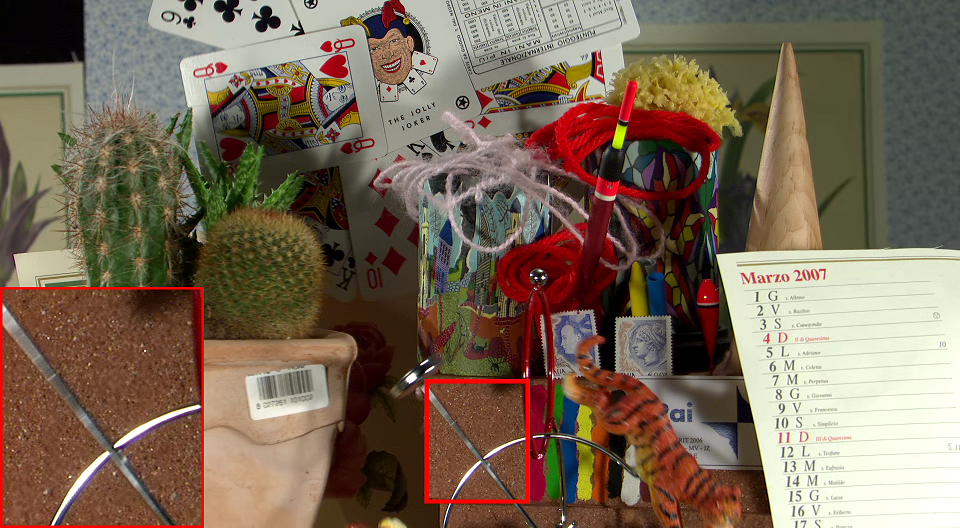} \\	
	\end{tabular}
	\caption{Cactus enhancement layer encoded with QP=22. The epitome size is 48.33\% of the input image. Before restoration, we can clearly notice in the red rectangle and in the calendar on the bottom right the blocks not belonging to the epitome from their blurry aspect. The E-LLM gives visually superior results compared to the E-LLE for the stochastic texture highlighted in the red rectangle. Both methods improve the quality of the straight lines in the calendar.}
	\label{fig:Cactus_example}
\end{figure*}

We give in Figs. \ref{fig:City_example} and \ref{fig:Cactus_example} visual examples of the enhancement layers for the City and Cactus images.
Note that these examples were chosen for their visual clarity, and do not necessarily correspond to the best RD performances.
In order to demonstrate the relevance of the post-processing step, we show on the top row the epitome EL before restoration, and on the bottom row the corresponding EL after applying the E-LLE and E-LLM methods.
Before restoration, the blocks not belonging to the epitome are particularly visible, as they are directly copied from the up-sampled decoded BL, and clearly lack high frequency details.
An obvious improvement of the quality can be observed after restoration for the high-frequency pseudo-periodic textures, such as the building of Fig. \ref{fig:City_example}, or the calendar of Fig. \ref{fig:Cactus_example}.
Although the E-LLM usually yields lower PSNR than the E-LLE method, we can see that it can perform visually better on high-frequency stochastic textures such as in the highlighted red rectangle of Fig. \ref{fig:Cactus_example}.

\subsection{Elements of complexity}

We give in this section some indications about the complexity, evaluated as the running time of the proposed methods.
We evaluate the complexity depending on the epitome size and input image size.
The results are averaged for each image class (which corresponds to an image size, see Table \ref{tab:test_im}).
Note that the epitome generation algorithm was implemented in C++, while the restoration methods were implemented in Matlab.

\begin{figure}[t]
	\centering
	\includegraphics[width=9cm,trim={1.6cm 2.2cm 2cm 1.9cm},clip]{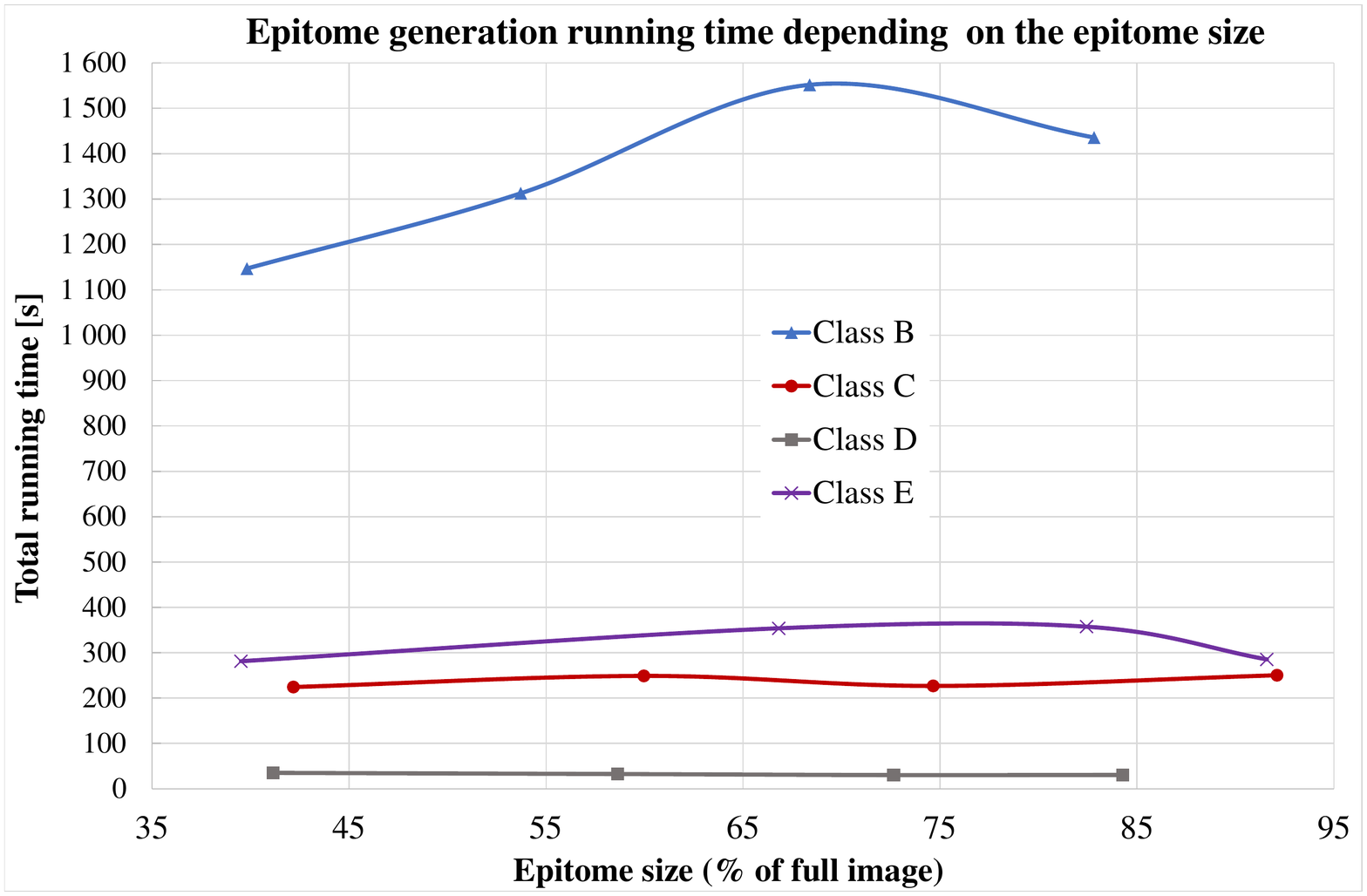}
	\caption{Epitome generation running time depending on the epitome size for different image classes.}
	\label{fig:time_eptm}
\end{figure}

We give in Fig. \ref{fig:time_eptm} the complexity of the epitome generation.
We can see that the epitome generation running time mainly varies depending on the input image size, while the epitome size has a limited impact.
For the biggest epitomes, which correspond to the best RD results, we observed that in average 50\% to 90\% of the complexity is dedicated to the self-similarities search step.
As this step is highly parallelizable, the total running time could be reduced by using a parallel implementation, e.g. on GPU.

\begin{figure}[ht]
	\centering
	\includegraphics[width=9cm,trim={1.6cm 2.2cm 2cm 1.9cm},clip]{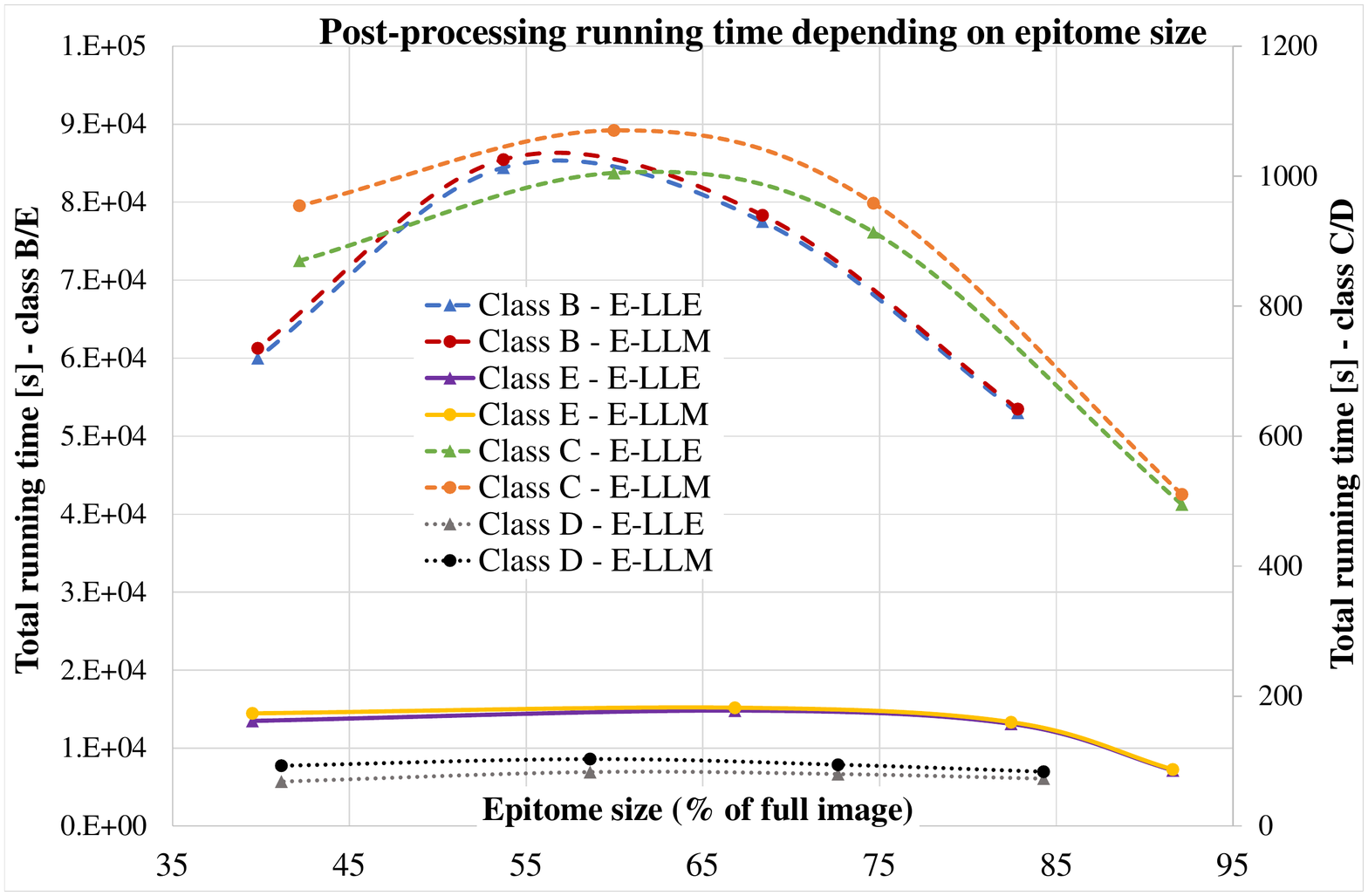}
	\caption{Post-processing running time of the different restoration methods depending on the epitome size for different image classes.}
	\label{fig:time_proc}
\end{figure}

We show in Fig. \ref{fig:time_proc} the complexity of the post-processing step. 
The processing time is similar for both E-LLE and E-LLM methods, and obviously increases with the size of the image.
However, we can observe that overall the complexity is reduced for the biggest epitomes, which interestingly corresponds to the best RD performances.
In fact, when transmitting bigger epitomes, less patches not belonging to the epitome have to be processed.

The simulations showed that in average, about 95\% of the post-processing complexity is dedicated to the $K$-NN search.
The $K$-NN search was performed with the Matlab \textit{knnsearch} function, based on $Kd$-tree \cite{Bentley1975,Freidman1977a}.
In order to reduce the total running time, the complexity of the $K$-NN search could be further reduced by using more advanced approximate nearest neighbor search \cite{Barnes2010,Dowson2011,Cherian2014,Zhu2014} or parallel implementation, possibly on GPU \cite{Garcia2008a}.

\subsection{Extension to scalable video coding}

The work presented in this paper is dedicated to scalable single image coding, however a straight extension to scalable video coding can be considered by applying the proposed method to each frame of the sequence.
Preliminary experiments are conducted on a set of 3 test sequences, consisting of 9 frames with a CIF resolution in order to limit the computation time.
The epitomes are generated using one matching threshold value $\varepsilon_M=7.0$.
In order to exploit the temporal redundancies, the $K$-NN search step is performed in the epitomes of the two closest frames in addition to the current one.

We show the RD performances measured with the Bjontegaard rate gains in Table \ref{tab:RD_perf_seq}.
These preliminary results indicate that the proposed scheme is also expected to bring significant bit-rate reduction when extended to full video sequences.
These results are not obvious to predict, since the inter layer prediction is here also competing with inter frame prediction modes, which are much more efficient than the intra prediction modes.

\begin{table}[ht]
	\footnotesize
  \centering
  \caption{Bjontegaard rate gains against SHVC}
    \begin{tabular}{|l|c|c|c|}
    \hline
    Sequence		& Epitome size (\%)  			&  \multicolumn{2}{c|}{BD rate gains (\%)} \\
    				& (averaged over all frames)	& E-LLE 	& E-LLM \\
    \hline
    City			& 56.66							& -26.56	& -26.59	\\
	Macleans		& 79.93							& -2.24		& -2.15		\\
	Mobile			& 82.22							& -10.12	& -10.20	\\
	\hline
    \end{tabular}%
  \label{tab:RD_perf_seq}%
\end{table}%

\section{Conclusion and future work}

We propose in this paper a novel scheme for scalable image coding, based on an epitomic factored representation of the enhancement layer and appropriate restoration methods at the decoder side.
Significant bit-rate reduction is achieved for the spatial scalable application when compared to the SHVC reference EL.
These achievements were possible because of the specific epitomic model we used, which provides relevant texture information and is especially suitable for scalable encoding.
Note that improvements of the standard tools have been recently proposed, such as the generalized inter-layer residual prediction \cite{Li2013,Laude2014,Aminlou2014}, or enhanced in-loop prediction mechanism for the EL \cite{Hoangvan2015,Hoangvan2016}.
The proposed approach is compatible with such improvement of scalable coding schemes, as the coding of the epitome as an enhancement layer would be improved as well.

The proposed scheme could be improved by studying alternative restoration approaches.
For instance, different regression could be considered for the E-LLM instead of the direct least-square approach, such as (Kernel) Ridge Regression \cite{Bevilacqua2014,Kim2010}. 
Alternatively, the restoration step at the decoder side could be considered as an inpainting problem with prior knowledge on the ``holes'' to be filled in the form of low resolution patches.
Inpainting has been extensively studied over the last decades (see \cite{Guillemot2014} and reference therein for more details), and the exemplar-based multi-scale approaches \cite{Drori2003,LeMeur2013} are well suited in our context.

Future work also includes the adaptation of the proposed scheme for scalable video coding, as preliminary results indicate promising RD performances. 
The epitomes were here generated separately for each frame.
The RD performances would benefit from an epitomic model which takes into account the temporal redundancies.
Furthermore, the image self-similarities in the current epitome are found using a single block matching algorithm, while the application at the decoder side is based on multi-patches techniques.
New epitomic models have been designed to take into account the epitome application in the generation process.
For example, an epitome is proposed in \cite{Turkan2015} for multi-patches super-resolution, which showed that a more compact representation can be obtained for a similar image reconstruction quality.
Such model could thus be considered in the proposed scheme in order to improve the RD performances, at the cost of an increased complexity.
In addition, the distortion minimization criterion of Eq. \ref{eq:ext_crit} used for the epitome charts creation could be changed into a rate-distortion optimization criterion, as in \cite{Cherigui2011}.
Ideally, the distortion would be directly computed on the restored EL instead of the reconstructed image, and the rate directly evaluated as the EL rate.

Finally, the proposed scheme can be extended to other scalable applications, such as color gamut or LDR/HDR scalabilities.
Even though we use LLE in this paper for super-resolution, it has been proven efficient for many different applications such as de-noising \cite{Shen2005}, image prediction \cite{Turkan2012}, or inpainting \cite{Guillemot2014}.
We can thus expect the LLE-based restoration methods to be efficient for different scalable applications.

\begin{table*}[htbp]
 \footnotesize
  \centering
  \caption[caption]{Bjontegaard rate gains against SHVC depending on the epitome size. \\ \textit{(For each image, the best rate saving is indicated in bold.})}
    \begin{tabular}{|c|c|c||c|c|c|}
    \hline
    Epitome size 		& \multicolumn{2}{c||}{BD rate gains (\%)}	& Epitome size 			& \multicolumn{2}{c|}{BD rate gains (\%)} \\
    (\% of input image) & E-LLE & E-LLM	 		 				& (\% of input image)	& E-LLE & E-LLM \\
    \hline
    \multicolumn{3}{|c||}{BasketballDrive}	& \multicolumn{3}{c|}{BasketballDrill}\\
	\hline
    90.62 & \textbf{-20.07} & -19.61      	& 87.05 & \textbf{-6.52} & -5.50  \\
    64.10 & -15.94 			& -13.70      	& 59.94 & -2.82 		 & 1.08   \\
    49.33 & -13.66 			& -8.69       	& 42.63 & -1.62 		 & 4.44   \\
    32.34 & -9.70  			& -0.08       	& 28.53 & 3.18  		 & 13.08  \\
    \hline
    \multicolumn{3}{|c||}{Cactus}	 		& \multicolumn{3}{c|}{Keiba} \\
    \hline                               
    79.85 & \textbf{-18.19} & -16.46 	    & 93.59 & \textbf{-6.71}	& -6.42 \\
    71.24 & -17.67 			& -15.14   	    & 81.28 & -3.69 			& -1.99  \\
    60.66 & -16.33 			& -13.01   	    & 63.53 & 3.24  			& 7.75   \\
    48.33 & -11.63 			& -7.42    	    & 40.77 & 16.06 			& 23.52  \\
	\hline                               
    \multicolumn{3}{|c||}{Ducks}	     	& \multicolumn{3}{c|}{Mall} \\
    \hline                               
    89.63 & \textbf{-19.52} & -19.07 	    & 92.95 & \textbf{-18.20}	& -16.76 \\
    77.41 & -16.71 			& -14.21 	    & 76.28 & -0.50  			& -2.13 \\
    48.28 & 2.88   			& 10.28  	    & 66.15 & -4.13  			& 3.04  \\
          &        			&        	    & 50.26 & 6.75   			& 27.54 \\
	\hline
    \multicolumn{3}{|c||}{Kimono} 	    	& \multicolumn{3}{c|}{PartyScene} \\
	\hline
    90.13 & \textbf{-21.75} & -21.42		& 94.82 & \textbf{-5.44} 	& -4.29 \\
    75.53 & -15.98 			& -12.63      	& 81.12 & -1.18 			& 7.20  \\
    59.36 & -17.37 			& -15.08      	& 67.56 & 8.83  			& 25.96 \\
    35.34 & -15.82 			& -12.31      	& 49.13 & 26.89 			& 57.15 \\
	\hline
    \multicolumn{3}{|c||}{ParkScene}	    & \multicolumn{3}{c|}{BasketballPass} \\
	\hline
    86.58 & \textbf{-16.88} & -16.45 		& 77.76 & \textbf{-16.17} 	& -13.15 \\
    73.55 & -15.10 			& -13.84  		& 66.60 & -14.07 			& -7.25 \\
    61.99 & -10.69 			& -7.50  		& 56.41 & -0.53  			& 10.48 \\
    47.18 & -3.94  			& 2.89   		& 42.31 & 5.72   			&  28.21 \\
	\hline
    \multicolumn{3}{|c||}{Tennis}		    & \multicolumn{3}{c|}{BlowingBubbles} \\
	\hline
    64.49 & \textbf{-23.04} & -21.91     	& 87.56 & \textbf{-6.33} 	& -3.29  \\
    50.44 & -22.03 			& -19.61     	& 73.33 & -2.73 			& 3.65   \\
    43.12 & -19.90 			& -16.59     	& 58.85 & 3.27  			& 13.95  \\
    32.22 & -18.42 			& -13.13     	& 36.92 & 16.81 			& 44.03  \\
	\hline    
    \multicolumn{3}{|c||}{Terrace} 	    	& \multicolumn{3}{c|}{RaceHorses}	\\
	\hline
    78.46 & \textbf{-13.27} & -12.49        & 91.03 & \textbf{-16.08} 	& -15.67 \\
    66.39 & -11.32 			& -9.57         & 79.23 & -4.49 			& -3.03  \\
    53.31 & -6.81  			& -3.01         & 58.14 & 6.69  			& 20.64  \\
    43.50 & -0.50  			& 9.03          & 36.67 & 23.45 			& 63.23  \\
	\hline
    \multicolumn{3}{|c||}{City}		    	& \multicolumn{3}{c|}{Square}	\\
	\hline
    91.59 & \textbf{-10.05} & -8.76			& 80.77  & \textbf{-6.45} 	& -2.97 \\
    82.44 & -6.24 			& -1.59  		&  71.41 & -5.88 			& 1.08  \\
    66.81 & 3.27  			& 17.75  		&  61.09 & -2.00 			& 4.75  \\
    39.52 & 28.00 			& 59.96  		&  48.72 & 9.80				& 31.68 \\
	\hline
    \end{tabular}%
  \label{tab:RD_perf}%
\end{table*}

\bibliographystyle{IEEEtran}
\bibliography{lib_filtered}

\end{document}